\documentclass{IEEEtran}
\usepackage{cite}
\usepackage{amsmath,amssymb,amsfonts}
\usepackage{algorithmic}
\usepackage{graphicx}
\usepackage{textcomp}

\usepackage[colorlinks,urlcolor=blue,linkcolor=blue,citecolor=blue]{hyperref}
\usepackage{color,array}
\usepackage{graphicx}
\usepackage{graphics} 
\usepackage{subcaption}
\newtheorem{defi}{Definition}[section]
\newtheorem{lem}{Lemma}[section]
\newtheorem{prop}{Proposition}[section]
\newtheorem{thm}{Theorem}[section]

\def\BibTeX{{\rm B\kern-.05em{\sc i\kern-.025em b}\kern-.08em
    T\kern-.1667em\lower.7ex\hbox{E}\kern-.125emX}}
\begin{document}
\title{Distributionally Robust Safety Verification of Neural Networks via Worst‑Case CVaR} 
\author{Masako Kishida, \IEEEmembership{Senior Member, IEEE}
\thanks{This work was supported by JST, PRESTO Grant Number JPMJPR22C3, Japan. }
\thanks{A preliminary version of this paper appears in the proceedings of IEEE Conference on Decision and Control 2025 \cite{Kis25}.} 
\thanks{The author is with 
National Institute of Informatics, Tokyo Japan (e-mail: kishida@ieee.org).}
\thanks{ChatGPT was used to improve writing.}
}

\maketitle

\begin{abstract}
Ensuring the safety of neural networks under input uncertainty is a fundamental challenge in safety-critical applications.
This paper builds on and expands Fazlyab’s quadratic-constraint (QC) and semidefinite-programming (SDP) framework for neural network verification to a distributionally robust and tail-risk-aware setting by integrating worst-case Conditional Value-at-Risk (WC-CVaR) over a moment-based ambiguity set with fixed mean and covariance.
The resulting conditions remain SDP-checkable and explicitly account for tail risk.
This integration broadens input-uncertainty geometry—covering ellipsoids, polytopes, and hyperplanes—and extends applicability to safety-critical domains where tail-event severity matters.
Applications to closed‑loop reachability of control systems and classification are demonstrated through numerical experiments, illustrating how the risk level $\varepsilon$ trades conservatism for tolerance to tail events—while preserving the computational structure of prior QC/SDP methods for neural network verification and robustness analysis.

\end{abstract}

\begin{IEEEkeywords}
Distributionally robust optimization, conditional value‑at‑risk (CVaR), quadratic constraints, semidefinite programming, reachability, neural network verification, tail risk.
\end{IEEEkeywords}

%%%%%%%%%%%%%%%%%%%%%%%%%%%%%%%%%%%%%%%%%%%%%%%%%%%%%%%%%%%%%%%%%%%%%%%%%%%%%%%%
\section{INTRODUCTION}

% talk about risk
Risk analysis and management are crucial in the development and operation of safety-critical systems such as automotive systems \cite{DeGES21}, avionics systems \cite{BruDA21}, and medical devices \cite{AleLK13}. 
These systems require rigorous safety guarantees despite inherent uncertainties, as their failure can lead to catastrophic consequences, including loss of life, substantial property damage, or environmental harm. 
Major approaches to handling uncertainties generally fall into two types: deterministic set-based approaches that provide worst-case guarantees \cite{ZhoDG96, DulP13, BlaM15} and probabilistic/stochastic approaches that explicitly represent uncertainty using probability distributions and impose performance goals using such as chance constraints  \cite{NemS07} or expected values \cite{WelB95}. 
While effective in many settings, both approaches may fail to capture rare but severe events---tail risks---that are unacceptable in safety-critical contexts.

% talk about neural network safety
In recent years, deep neural networks (DNNs) have become ubiquitous in control and decision-making systems. However, DNNs are known to be vulnerable to adversarial perturbations \cite{GooSS14,BigR18, MadMS17}. 
This has motivated extensive research on verification and robustness analysis of DNNs \cite{KatBD17,Ehl17,BunTT20,ShiZC20, XuZW21,  FazMP22,FazMP21, FazMP19, NooHD24, SalYZ19, ChiCZ25, WenCN19, PilSO23, HuaKW17}. 
Existing DNN verification approaches can be broadly categorized into three classes:
\begin{itemize}
  \item[(i)] Exact methods, such as SMT/MILP solvers~\cite{KatBD17,Ehl17} 
  and branch-and-bound algorithms~\cite{BunTT20}. 
  They provide sound and complete guarantees.
  
  \item[(ii)] Bound-propagation methods, 
  including abstract interpretation and forward–backward propagation~\cite{ShiZC20}, 
as well as recent frameworks that use the backward mode linear relaxation based perturbation analysis~\cite{XuZW21}. 
 These typically yield conservative bounds.
 
  \item[(iii)] Convex optimization methods, 
  e.g., quadratic-constraint and semidefinite-programming (QC/SDP)-based analysis~\cite{FazMP22,FazMP21, FazMP19} 
  and control-theoretic stability analysis of recurrent networks~\cite{NooHD24}. 
  They offer elegant convex formulations and strong theoretical foundations~\cite{SalYZ19}.
  Recent work also explores hybrid bound-propagation–SDP approaches that inject SDP-derived inter-neuron coupling
  into scalable bound propagation~\cite{ChiCZ25}.
\end{itemize}
While these approaches have advanced verification significantly, 
they do not explicitly quantify the severity of low-probability catastrophic failures.

\begin{table*}[ht]
    \centering
\caption{Comparison of three QC/SDP-based neural network verification approaches.}
    \label{tab:comparison}
\renewcommand{\arraystretch}{1.3}
\begin{tabular}{|p{3cm}|p{4cm}|p{4cm}|p{4cm}|}
\hline
\textbf{Category} &  \textbf{Fazlyab et al., 2022 \cite{FazMP22}} & \textbf{Fazlyab et al., 2019 \cite{FazMP19}} &\textbf{This paper} \\
\hline
\textbf{Approach} 
& Norm bounded 
& Confidence ellipsoid propagation  
& Assessment by Worst-Case Conditional Value-at-Risk (WC-CVaR)\\
\hline
\textbf{Uncertainty} 
& Bounded sets 
& Gaussian as well as random vectors with known mean and covariance 
& Distributional uncertainty: ambiguity set with known mean and covariance, tail risk considered \\
\hline
\textbf{Input geometry} 
& Those that can be expressed using affine or quadratic functions 
& Ellipsoids (use Chebyshev's inequality) 
& Those that can be expressed using affine or quadratic functions \\
\hline
\textbf{Output geometry} 
& Those that can be expressed using affine or quadratic functions 
& Ellipsoids 
& Those that can be expressed using affine or quadratic functions\\
\hline
\textbf{Metric} 
& Worst-case safety (all admissible inputs must satisfy specification) 
&  $p$-level confidence region
& WC-CVaR, explicitly accounting for tail risk \\
\hline
\end{tabular}
\end{table*}

In this paper, we extend Fazlyab's QC/SDP methods in (iii) by incorporating the Worst-Case Conditional Value-at-Risk (WC-CVaR) \cite{ZymKR13-b}, 
thereby enabling tail-risk-aware safety verification under distributional uncertainty.
Originally developed in finance \cite{RocU00,ZhuF09}, CVaR and its worst-case variant have recently been applied in control \cite{CleLI22,Kis24,KisC23} and machine learning \cite{TakS08,HirIM19,SomY20}. 
WC-CVaR quantifies the expected loss in the worst $\varepsilon$-tail over all distributions sharing prescribed mean and covariance. 
This distributional viewpoint is closely related to the literature on distributionally robust optimization (DRO), where ambiguity sets defined by moment information or other statistics are used to model distributional uncertainty \cite{DelY10,WieKS14}. 
It avoids reliance on a specific model (e.g., Gaussian) and is attractive because (a) moments are often available from data, and (b) the induced verification conditions remain SDP-checkable via QCs. 
Incorporating WC-CVaR provides a systematic trade-off between conservatism and tail-risk tolerance, while preserving SDP tractability. 
It also generalizes input-uncertainty descriptions beyond ellipsoids \cite{FazMP19} to polytopes and hyperplanes, broadening applicability (Table~\ref{tab:comparison}).

The contributions of this paper are as follows:
\begin{itemize}
    \item We extend Fazlyab's QC/SDP methods by incorporating WC-CVaR, yielding a distributionally robust method for neural network safety verification.  
    \item We establish a connection between WC-CVaR and confidence ellipsoid methods \cite{FazMP19} (Lemma~\ref{lem:eq}), showing that the two safe sets coincide when those sets are restricted to a special ellipsoid.  
    \item We illustrate the proposed methods through numerical experiments on closed-loop reachability and classification problems, highlighting performance under both light- and heavy-tailed input distributions.  
\end{itemize}

The paper is organized as follows: Section \ref{sec:pre} introduces notation and WC-CVaR fundamentals. Section \ref{sec:main} presents our risk-aware neural network verification framework. Section \ref{sec:app} discusses control and classification applications with numerical experiments. Finally, Section \ref{sec:conc} concludes the paper.

%%%%%%%%%%%%%
\section{Preliminaries}\label{sec:pre}

%------------------
\subsection{Notation}
Let $\mathbb{R}$, $\mathbb{R}^n$, $\mathbb{R}^{n\times m}$, and $\mathbb{S}^n$ denote the sets of real numbers, $n$-dimensional real vectors,  $n \times m$ real matrices, and $n \times n$ symmetric matrices, respectively. 
For $v \in \mathbb{R}^n$, let $\|v\|$ be its Euclidean norm, and
for $M \in \mathbb{R}^{n\times m}$, let $M^\top$ be its transpose and $\text{Tr}(M)$ its trace. 
$M \succ 0$ indicates that $M \in \mathbb{S}^n$ is positive definite. 
The all-ones vector in $\mathbb{R}^n$ and the identity matrix in $\mathbb{R}^{n\times n}$ are denoted by $\mathbf{1}_n$ and $I_n$ the $n \times n$, respectively, and subscripts are dropped when the sizes are clear.
For $x \in \mathbb{R}$, define $(x)^+ := \max\{x, 0\}$.
$\mathbb{E}_{\mathbb{P}}[x]$ and $\operatorname{Cov}_{\mathbb{P}}[x]$ denote the mean and covariance of a random vector 
$x \sim \mathbb{P}$, subscripts are dropped when the underlying distribution is clear.

%------------------
\subsection{Worst-Case Conditional Value-at-Risk}
CVaR at risk level $\varepsilon$ is the expected value beyond a certain high percentile, measuring the tail risk. 
Since the exact distribution is often unknown, we consider an ambiguity set of possible distributions $\mathcal{P}$. WC-CVaR quantifies the worst-case tail risk at level $\varepsilon$ over $\mathcal{P}$. This conservative metric guarantees robustness against any distribution in $\mathcal{P}$, not just a nominal one. 

We first define the ambiguity set and augmented moment matrix.
Let $\xi \in \mathbb{R}^n$ be a random vector with distribution $\mathbb{P}$.  
The moment-based ambiguity set is defined as
\begin{align}
\mathcal{P}(\mu,\Sigma) := \left\{\, \mathbb{P} \;:\;
\mathbb{E}_{\mathbb{P}}[\xi] = \mu,\;
\mathrm{Cov}_{\mathbb{P}}[\xi] = \Sigma \,\right\},
\end{align}
i.e., all distributions that share the same mean $\mu$ and covariance $\Sigma$, including both light-tailed and heavy-tailed families. 

The augmented second-order moment matrix of the lifted variable
$\bar{\xi} := [\xi^\top,\, 1]^\top$ is
\begin{align}\label{eq:omega}
\Omega = 
\begin{bmatrix}
\Sigma + \mu \mu^\top & \mu \\
\mu^\top & 1
\end{bmatrix},
\end{align}
which compactly represents the second-order information of $\bar{\xi}$.  
For brevity, we write $\mathcal{P}$ for $\mathcal{P}(\mu,\Sigma)$.

\begin{defi}[WC-CVaR \cite{ZymKR13-b}]
For a measurable loss function $L: \mathbb{R}^n \rightarrow \mathbb{R}$ and a risk level $\varepsilon \in (0,1)$, the Worst-Case Conditional Value-at-Risk (WC-CVaR) over the ambiguity set $\mathcal{P}$ is defined as:
 \begin{align}
\sup_{\mathbb{P}\in \mathcal{P}}\mathbb{P}\text{-CVaR}_{\varepsilon}[L({\xi})]
:=\inf_{\beta \in \mathbb{R}} \left\{ \beta + \frac{1}{\varepsilon}\sup_{\mathbb{P}\in \mathcal{P}}\mathbb{E}_{\mathbb{P}}[(L({\xi})-\beta)^+]\right\}.
\end{align}
\end{defi}
WC-CVaR quantifies the tail risk under the worst-case probability distribution within $\mathcal{P}$.
In this paper, the risk level $\varepsilon$ serves as a tuning parameter, where a smaller $\varepsilon$ makes the constraint more conservative, providing a trade-off between risk aversion and performance.

WC-CVaR is one of the coherent risk measure and satisfies the monotonicity property, which we will utilize.
\begin{prop}[Monotonicity property \cite{ZhuF09,Art99}]\label{prop:sub}
For measurable loss functions $L_1(\xi)$ and $L_2(\xi)$, if $L_1(\xi) \leq L_2(\xi)$ almost surely, then
\begin{align*}
\sup_{\mathbb{P}\in \mathcal{P}}\mathbb{P}\text{-CVaR}_{\varepsilon}[L_1(\xi)] \leq
\sup_{\mathbb{P}\in \mathcal{P}}\mathbb{P}\text{-CVaR}_{\varepsilon}[L_2(\xi)].
\end{align*}
\end{prop}

Other coherent risk measures could extend Fazlyab’s framework, but WC-CVaR offers superior computational tractability through Proposition \ref{prop:CVaR_SDP}.

\begin{prop}[Quadratic Loss Function \cite{ZymKR13-b}, \cite{ZymKR13}] \label{prop:CVaR_SDP}
For $L({\xi}) = \xi^\top \Pi \xi + 2\theta^\top \xi + \rho$, where $\Pi \in \mathbb{S}^n$, $\theta \in \mathbb{R}^n$, and $\rho\in \mathbb{R}$, the WC-CVaR is given by:
 \begin{align*}
\sup_{\mathbb{P}\in \mathcal{P}}\mathbb{P}\text{-CVaR}_{\varepsilon}[L({\xi})] =
&\inf_{ \beta} \left\{\beta + \frac{1}{\varepsilon}\text{Tr}(\Omega N): \right.\\
& N \succcurlyeq 0, \
\left. N-\left[\begin{array}{cc}\Pi &\theta \\ \theta^\top & \rho-\beta \end{array}\right] \succcurlyeq 0\right\}.
\end{align*}
\end{prop}

%%%%%%%%%%%%%
\section{Risk-Aware Neural Network Verification} \label{sec:main} 
This section proposes risk-aware DNN safety verification method that incorporates WC-CVaR.
We also discuss its complexity and extensions. 
%------------------------------------------
\subsection{Preparations} \label{sec:setup} 
In the rest of this paper, let $x  \sim \mathbb{P}\in \mathcal{P}(\mu, \Sigma)$ be a random vector of length $n$.  The augmented second-order moment matrix of $[x^\top, 1]^\top$ is given by \eqref{eq:omega}.

For safety verification, we utilize QCs, which were developed in robust control \cite{MegR02}.
In particular, we extend the approaches by Fazlyab et al. \cite{FazMP19,FazMP21, FazMP22} to account for tail risks as follows.
\begin{defi}[Risk-aware QCs] \label{def:QC}
For a symmetric, possibly indefinite matrix $H \in \mathbb{S}^{n+1}$, we define a risk-aware QC (with WC-CVaR risk level $\varepsilon$) by
\begin{align}\label{eq:QC}
\sup_{\mathbb{P}\in \mathcal{P}}\mathbb{P}\text{-CVaR}_{\varepsilon}\left[\begin{bmatrix}x\\ 1\end{bmatrix}^\top H \begin{bmatrix}x\\ 1\end{bmatrix}\right]  \leq 0.
\end{align}
Let $\mathcal{H}_{x} \subset \mathbb{S}^{n+1}$ be a set of $H$ that satisfies \eqref{eq:QC}.
Then, we say the ambiguity set $\mathcal{P}$ satisfies the QC defined by $\mathcal{H}_{x}$. 
\end{defi}

This means that, for most $x \sim \mathbb{P}\in \mathcal{P} $, it holds that
\begin{align} \label{eq:QC2}
\begin{bmatrix}x\\ 1\end{bmatrix}^\top H\begin{bmatrix}x\\ 1\end{bmatrix} \leq 0,
\end{align}
and even the mean of the worst $ 100\varepsilon$ \% of $x$ satisfies \eqref{eq:QC2}.

Here, we note the condition \eqref{eq:QC} can be checked by solving a SDP using Proposition \ref{prop:CVaR_SDP}.

%------------------------------------------
\subsection{Neural Network Model}

We consider an $\ell$-hidden-layer feed-forward fully connected neural network $f:\mathbb{R}^n \to \mathbb{R}^m$ defined by
\begin{align}\begin{aligned} \label{eq:NN}
x^0 & = x,\\
x^{k+1} &=  \phi(W^kx^k+b^k), \ k = 0, \dots, \ell-1, \\
f(x) &=W^{\ell}x^{\ell}+b^{\ell},
\end{aligned}\end{align}
where $x^0 = x \in \mathbb{R}^n$ is the input, 
$x^k \in \mathbb{R}^{n_k}$ is the output of the $k$th layer, 
$W^k \in \mathbb{R}^{n_{k+1}\times n_k}$ and $b^k \in \mathbb{R}^{n_{k+1}}$ 
are the weight matrix and bias vector for the $(k+1)$th layer, 
and $W^\ell \in \mathbb{R}^{m\times n_\ell},\ b^\ell \in \mathbb{R}^m$ 
define the final affine map.

The activation function $\phi$ is applied element-wise:
\begin{align}
\phi(x) =\begin{bmatrix}\varphi(x_1) &\dots & \varphi(x_{n_k}) \end{bmatrix}^\top, \ x \in \mathbb{R}^{n_k}.
\end{align}
Assuming identical activation functions across layers, the model \eqref{eq:NN} can be rewritten in a compact form:
\begin{align}\begin{aligned} \label{eq:compact}
x &= \mathbf{E}^0\mathbf{x}, \\
 \mathbf{B}\mathbf{x} &= \phi(\mathbf{A}\mathbf{x} + \mathbf{b}), \\
  f(x) &= W^\ell\mathbf{E}^\ell\mathbf{x} + b^\ell, 
\end{aligned}\end{align}
using the entry selector matrices $\mathbf{E}^k	\in \mathbb{R}^{n_k \times \sum_{i=0}^{\ell}n_i}$ and
\begin{align}\begin{aligned} \label{eq:selector}
\mathbf{x} &= [x^{0\top} \cdots x^{\ell\top}]^\top \in \mathbb{R}^{\sum_{i=0}^{\ell} n_i} , \\ 
x^k &= \mathbf{E}^k\mathbf{x},\ k = 0,\ldots,\ell,\\
\mathbf{A} &= \begin{bmatrix}
\text{diag}(W^0, W^1, \ldots, W^{\ell-1})& 0\\
\end{bmatrix}, \\
\mathbf{b}  &= \begin{bmatrix}
b^{0^\top} &
b^{1^\top}&
\cdots &
b^{{\ell-1}^\top} 
\end{bmatrix}^\top
,\\
\mathbf{B} &= \begin{bmatrix}
0 & I_{\sum_{i=1}^{\ell}n_i}
\end{bmatrix}.
\end{aligned}\end{align}

%------------------------------------------
\subsection{Risk-Aware QC for Input}
The  input $x \sim \mathbb{P} \in \mathcal{P}$ in \eqref{eq:compact} is characterized using the set $\mathcal{H}_x$, which contains symmetric indefinite matrices $H = -P$ that satisfy the following risk-aware QC \footnote{The negative sign follows the convention in \cite{FazMP19,FazMP21,FazMP22}.}:
\begin{align} \label{eq:input}
\sup_{\mathbb{P} \in \mathcal{P}} \mathbb{P}\text{-CVaR}_{\varepsilon} \left[-\begin{bmatrix} x \\ 1 \end{bmatrix}^\top P \begin{bmatrix} x \\ 1 \end{bmatrix} \right] \leq 0.
\end{align}
If \eqref{eq:input} holds, then we say $\mathcal{P}$ satisfies the risk-aware QC defined by $\mathcal{H}_x$.

This QC encodes the risk-aware safety verification's assumption that the inputs lie within a risk-bounded set.
%------------------------------------------
\subsection{QC for Activation Function}
Most standard activation functions, including sector-bounded, slope-restricted, and bounded functions, can be described using QCs \cite{FazMP19,FazMP21,FazMP22}. Because the activation function we consider itself is deterministic, we have QCs instead of risk-aware QCs for the activation functions.

Let $\phi: \mathbb{R}^d \to \mathbb{R}^d$, and suppose $\mathcal{Q}_{\phi}\subset \mathcal{S}^{2d+1}$ is the set of symmetric indefinite matrices $Q$ such that the inequality
\begin{align} \label{eq:Q}
\begin{bmatrix}
z \\
\phi(z) \\
1
\end{bmatrix}^{\top}
Q
\begin{bmatrix}
z \\
\phi(z) \\
1
\end{bmatrix} \geq 0
\end{align}
holds for all $z \in \mathbb{R}^d$. Then, we say $\phi$ satisfies QC defined by $\mathcal{Q}_{\phi}$.

Intuitively, this QC bounds the possible values of the hidden-layer signals $z$ and their activations $\phi(z)$, ensuring the nonlinearity does not produce extreme outputs outside a known sector.
 
Constructing a tight $\mathcal{Q}_{\phi}$ can be challenging. However, ReLU has well-established QCs using slope and repeated nonlinearity constraints, while Sigmoid and Tanh benefit from sector and local QCs for tighter bounds. Preprocessing techniques further improve accuracy by refining local activation bounds \cite{FazMP22}.
Once $\mathcal{Q}_{\phi}$ is fixed and $\phi$ satisfies the QC defined by $\mathcal{Q}_{\phi}$, we can utilize \eqref{eq:Q} to bound the behavior of activation functions. See Appendix \ref{appendix:Q} for the exact expression of $Q$  in the case of ReLU.

%------------------------------------------
\subsection{Risk-Aware QC for Safety Verification}
The risk-aware safe set for the output $f(x)$ of the neural network \eqref{eq:compact}  can be formulated similarly to the input constraint using a symmetric indefinite matrix $S$. Specifically, we consider the following risk-aware QC:
\begin{align} \label{eq:output}
\sup_{\mathbb{P} \in \mathcal{P}} \mathbb{P}\text{-CVaR}_{\epsilon} \left[\begin{bmatrix}
x \\ f(x) \\ 1
\end{bmatrix}^{\top} S
\begin{bmatrix}
x \\ f(x) \\ 1
\end{bmatrix} \right] \leq 0.
\end{align}
If \eqref{eq:output} holds, then we say $f(x)$ satisfies the risk-aware QC defined by $S$.

This QC ensures that the network’s final-layer output remains within a set constrained by a risk-aware bound.

%------------------------------------------
\subsection{Sufficient Risk-aware Safety Condition}
With the neural network model \eqref{eq:compact} and QCs \eqref{eq:input}-\eqref{eq:output} defined, we now present the main  result. The following theorem states sufficient conditions for verifying that a neural network meets the risk-aware safety specification \eqref{eq:output} under input uncertainties \eqref{eq:input}.

\begin{thm} \label{thm:sufficient}
Consider the neural network \eqref{eq:compact}.
Suppose that $\mathcal{H}_x$ and $\mathcal{Q}_{\phi}$  are given. 
Assume that $\mathcal{P}$ satisfies the risk-aware QC defined by $\mathcal{H}_x$ and $\phi(z)$ satisfies the QC defined by $\mathcal{Q}_{\phi}$. 
If a symmetric matrix $S$ satisfies the linear matrix inequality (LMI)
\begin{align}
M_{\text{in}}(P) + M_{\text{mid}}(Q) + M_{\text{out}}(S) \preceq 0, \label{eq:sum}
\end{align}
where
\begin{align}
\begin{aligned}
M_{\text{in}}(P) &=
\begin{bmatrix}
\mathbf{E}^0 & 0 \\
 0 & 1
\end{bmatrix}^\top P\begin{bmatrix}
\mathbf{E}^0 & 0 \\
 0 & 1
\end{bmatrix},  \\
M_{\text{mid}}(Q) &=
\begin{bmatrix}
\mathbf{A}  & \mathbf{b} \\
\mathbf{B} & 0 \\
0 & 1
\end{bmatrix}^\top Q\begin{bmatrix}
\mathbf{A}  &  \mathbf{b} \\
\mathbf{B}  & 0 \\
0 & 1
\end{bmatrix}, \\
M_{\text{out}}(S)& =
\begin{bmatrix}
\mathbf{E}^0& 0  \\
W^{\ell}\mathbf{E}^{\ell}&  \mathbf{b}^{\ell} \\
 0 & 1
\end{bmatrix}^\top S\begin{bmatrix}
\mathbf{E}^0& 0  \\
W^{\ell} \mathbf{E}^{\ell}&  \mathbf{b}^{\ell} \\
 0 & 1
\end{bmatrix}
\end{aligned}
\end{align}
for some $-P\in \mathcal{H}_x$, $Q \in Q_{\phi}$, then the output satisfies the risk-aware safety constraint \eqref{eq:output}.
\end{thm}
\begin{IEEEproof}
See Appendix \ref{proof}.
\end{IEEEproof}
Here, $S$ encodes a fixed safety specification to be checked, and the optimization variables are $P$ and $Q$. 
In practice, once the input set is specified, the matrix $P$ is typically fixed, 
and one may optimize over $Q$  to reduce conservatism in the output bound.

%------------------------------------------
\subsection{Computing a Minimum-volume Ellipsoidal Safe Set}

For a fixed neural network and input constraint with $P$, when the output variable for safety verification is defined as a linear map of the input and the output of the neural network,
\begin{align}\label{eq:y}
y = C \begin{bmatrix} x \\ f(x) \end{bmatrix} \in \mathbb{R}^{m_y},
\end{align}
a minimum volume ellipsoidal safe set for $y$, s.t. 
\begin{align}
\sup_{\mathbb{P} \in \mathcal{P}} \mathbb{P}\text{-CVaR}_{\varepsilon} \left[y^\top E^{-1} y -1\right] \leq 0.
\end{align}
 with $E \succ 0$,  can be obtained via convex optimization
\begin{align}\begin{aligned}\label{eq:ell}
&\min_{E \succ 0} 
\; -\log\det(E^{-1}) 
\\
& \text{s.t.} \quad
M_{\mathrm{in}}(P) + M_{\mathrm{mid}}(Q) + M_{\mathrm{out}}(S(E)) \preceq 0,
\end{aligned}\end{align}
where
\begin{align}
S(E) = \begin{bmatrix}
C^\top E^{-1} C & 0 \\
0 & -1
\end{bmatrix}.
\end{align}

%------------------------------------------
\subsection{Choosing $\varepsilon$} 
In safety critical applications, $\varepsilon$ should reflect the tolerated tail after accounting for dataset size and uncertainty in 
$\mu$ and $\Sigma$. A smaller $\varepsilon$ tightens certification but may increase conservatism. We recommend validating with empirical CVaR on held‑out data at the certified $\varepsilon$.

%------------------------------------------
\subsection{Complexity and scalability}

The neural network \eqref{eq:compact} has $\ell+1$ layers with widths $\{n_k\}_{k=0}^{\ell}$ ($n=n_0$ input, $m=n_{\ell+1}$ output).

The LMI condition \eqref{eq:sum} in Theorem~\ref{thm:sufficient} checks a fixed safety specification $S$,
which in practice is solved as an SDP feasibility problem.
The optimization variables are the multipliers $-P \in \mathcal{H}_x$ and $Q \in \mathcal{Q}_\phi$, 
while $S$ is given. The aggregated variable  (in Appendix \ref{proof}) is 
\begin{align}
\bar x := [x_0^\top, x_1^\top, \dots, x_{\ell}^\top, 1]^\top \in \mathbb{R}^{\bar n}
\end{align}
and the lifted dimension is 
\begin{align}
\bar{n} = \sum_{k=0}^\ell n_k+ 1 .
\end{align}
The number of variables in $Q$ grows as 
$O(\sum_{k=1}^{\ell -1} n_k)$ for diagonal multipliers 
and $O(\sum_{k=1}^{\ell -1} n_k^2)$ for repeated-nonlinearity multipliers.

The SDP in \eqref{eq:ell} searches for an ellipsoidal safe set. 
Here $S$ is parameterized by a matrix $E \succ 0$ of size $m_y \times m_y$, 
and the optimization variables are $Q$ and $E$ (with $P$ fixed in practice). 
The matrix $E$ adds $m_y(m_y+1)/2$ scalar variables to the problem, 
on top of those from the activation multipliers in $Q$. 

Both SDP use LMIs of size $\bar{n}$, 
so the practical bottleneck is the scaling of the multipliers and the output dimension.

%------------------------------------------
\subsection{Extensions beyond standard feed-forward MLPs}\label{sec:extensions}

Although our discussions have focused on feed-forward multilayer perceptrons (MLPs), the proposed WC-CVaR + QC/SDP framework extends naturally to more general neural network architectures.

\begin{itemize}

\item \textbf{Residual neural network (ResNet):}
A residual block augments a linear transformation with an identity skip connection, e.g., $x^{k+1}=\phi(W^k x^k+b^k)+x^k$.
Thus, the verification conditions retain the same SDP form with $\mathbf{B}$ replaced by a block-difference operator; only $M_{\text{mid}}(Q)$ is modified.

\item \textbf{Recurrent Neural Networks (RNNs):}
An RNN takes the form $h^{k}=\phi(W x^{k}+U h^{k-1}+b)$ with hidden state $h^{k}$.
Over a finite horizon, time-unfolding preserves the SDP structure with a block-banded lifting.
Modify $M_{\text{mid}}(Q)$ by substituting $(\mathbf{A},\mathbf{b},\mathbf{B})$ with their recurrent counterparts, include $h^{0}$ in the input QC.

\end{itemize}

More generally, any architecture that can be expressed as linear operators composed with slope-/sector-bounded nonlinearities—possibly with skip connections, convolutions, graph propagations, or time unrolling—fits the same QC/SDP template by replacing $(\mathbf{A},\mathbf{b},\mathbf{B})$ accordingly.

These extensions demonstrate that the proposed risk-aware verification framework is not restricted to simple feed-forward MLPs, while preserving the tractable SDP structure of Theorem~\ref{thm:sufficient}.

%%%%%%%%%%%%%%%%%%%%%
\section{Applications}\label{sec:app}
Here we show how the obtained results can be used for closed-loop reachability and classification problems. 
Throughout this section, we restrict activation functions to ReLU.  
%--------------------
\subsection{Closed-loop Reachability}

Consider a discrete-time linear time-invariant system with a neural network controller, as in \cite{FazMP22}:
\begin{align}\label{eq:LTI}
x^{+} = A x + B f(x),
\end{align}
where $x \in \mathbb{R}^n$ is the current state, $A \in \mathbb{R}^{n \times n}$ and $B \in \mathbb{R}^{n \times m}$ are the system matrices, and $f: \mathbb{R}^n \to \mathbb{R}^m$ is a neural network controller with $x$ as its input.

Here, we consider obtaining a minimum volume ellipsoidal safe set for the output $y = x^+$, 
with input set being the set of possible current states $x$. 
In this case, $C$ in \eqref{eq:y} is
\begin{align}
C = 
 \begin{bmatrix}
A&B \end{bmatrix}.
\end{align}

Numerical experiments were performed with the system parameters
\begin{align}
A =   
 \begin{bmatrix}
0.2&0 \\
0.1 & 0.3
\end{bmatrix}, \ 
B = 
\begin{bmatrix}
-1\\0
\end{bmatrix}, 
\end{align}
and a neural network controller that approximates a stabilizing controller $f(x) \approx Kx$ with 
$
K = [
-1\ 2],
$ using 2 neurons in the input layer, 3 neurons in the hidden layer, and 1 neuron in the output layer. 

We compared three approaches:
\begin{enumerate}
\item Norm bounded  \cite{FazMP22},
\item Confidence level \cite{FazMP19},
\item Risk-aware (proposed)
\end{enumerate}
To illustrate sampled inputs and outputs, we used the followings:
\begin{itemize}
\item For 1), the inputs are random vectors $x$ drawn uniformly from the unit disk, with mean $\mu = 0$ and covariance $\Sigma = 1/4 I$. Those represents samples in the deterministic input region.
\item For 2) and 3), the inputs are random vectors $x$ drawn from uniform, normal and $t$-distributions, all with mean $\mu = 0$ and covariance $\Sigma = 1/4 I$.\end{itemize}
In each case, the outputs are computed by \eqref{eq:LTI}.

We set the input  QC with 
\begin{align}
P &=   
 \begin{bmatrix}
-I& 0 \\
0 & r\\
\end{bmatrix},
\end{align}
where 
\begin{itemize}
\item Case 1): $r=1$ 
\item Case 2): $
r =\dfrac{1}{2(1-p)}$
\item Case 3): $
r = \sup_{\mathbb{P} \in \mathcal{P}} \mathbb{P}\text{-CVaR}_{\epsilon}\begin{bmatrix}
x^2
\end{bmatrix}=\dfrac{1}{2\varepsilon}
$,
\end{itemize}
Although the chance-constrained \cite{FazMP19} and WC-CVaR formulation are derived from different risk measures, our case provides identical bounds.  

\begin{lem}\label{lem:eq}
Suppose the input random vector is drawn from a distribution with mean $\mu=0$ and covariance $\Sigma$.
If we fix the shape of input bound to be a scaled ellipsoid $\{x: x^T \Sigma^{-1} x \leq k\}$ with some $k>0$, then
the risk-aware input set with $\varepsilon$ is identical to the ellipsoidal chance-constraint input set  with $1-p$.
\end{lem}
\begin{IEEEproof}
 See appendix \ref{appendix:equivalence}.
\end{IEEEproof}
Furthermore, both of the corresponding minimum-volume ellipsoidal safe sets can be computed using \eqref{eq:ell}, thus coincide. 

For the confidence levels in 2), we plot $p=0.1, 0.5$ and 0.9 and for the risk levels in 3), we plot $\varepsilon=0.1, 0.5$ and 0.9.

\begin{figure}[tb]\vspace{.05in}
    \centering

    % (a)
    \begin{subfigure}{\linewidth}
        \centering
    \includegraphics[width=\linewidth, viewport=60 20 1045 530, clip]{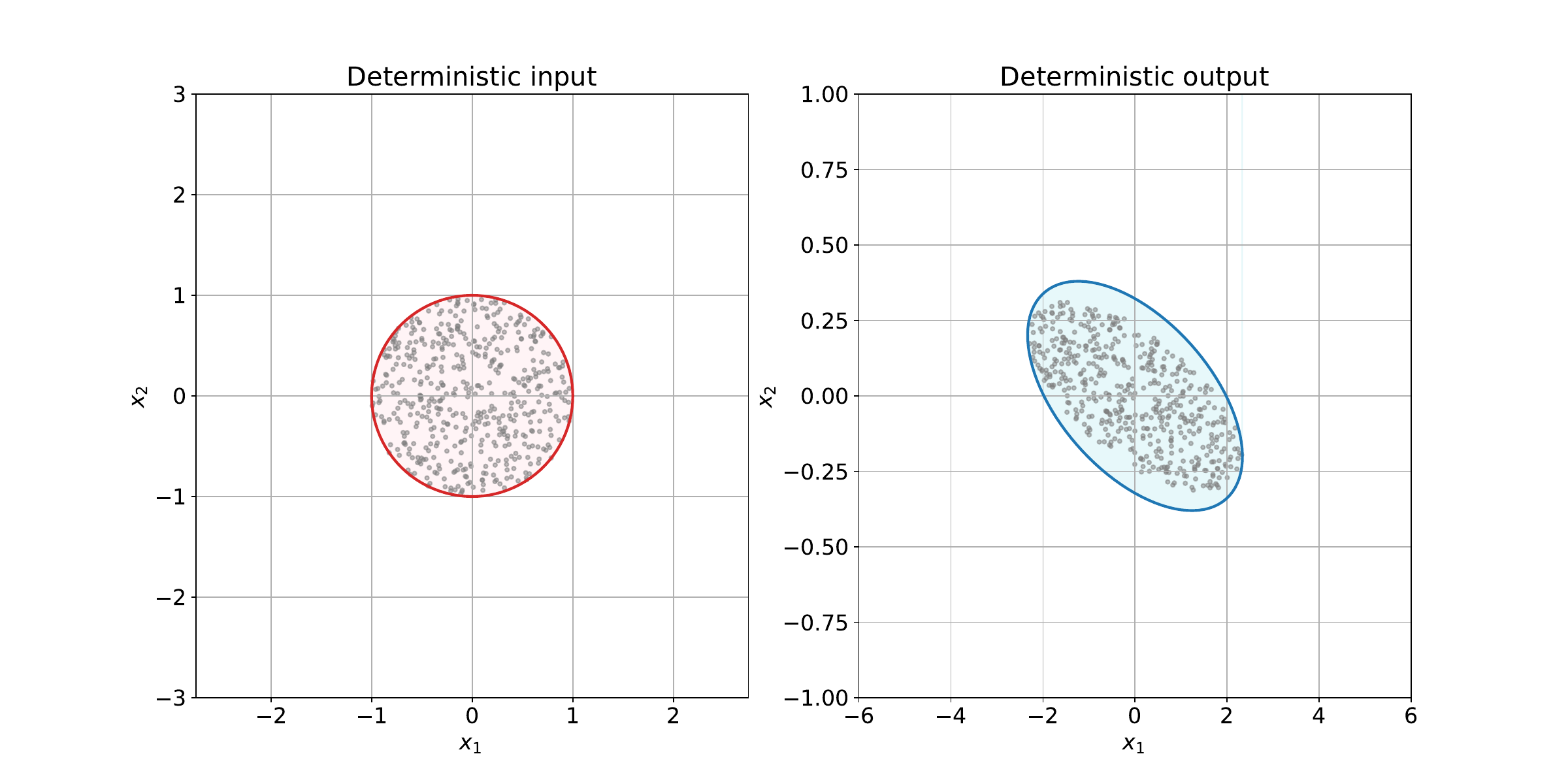}  \\
    \subcaption{Norm bounded \cite{FazMP22}}
        \label{fig:det}
    \end{subfigure}
    
    \vspace{5pt} %
    
    % (b)
    \begin{subfigure}{\linewidth}
        \centering
   \includegraphics[width=\linewidth, viewport=60 20 1045 530, clip]{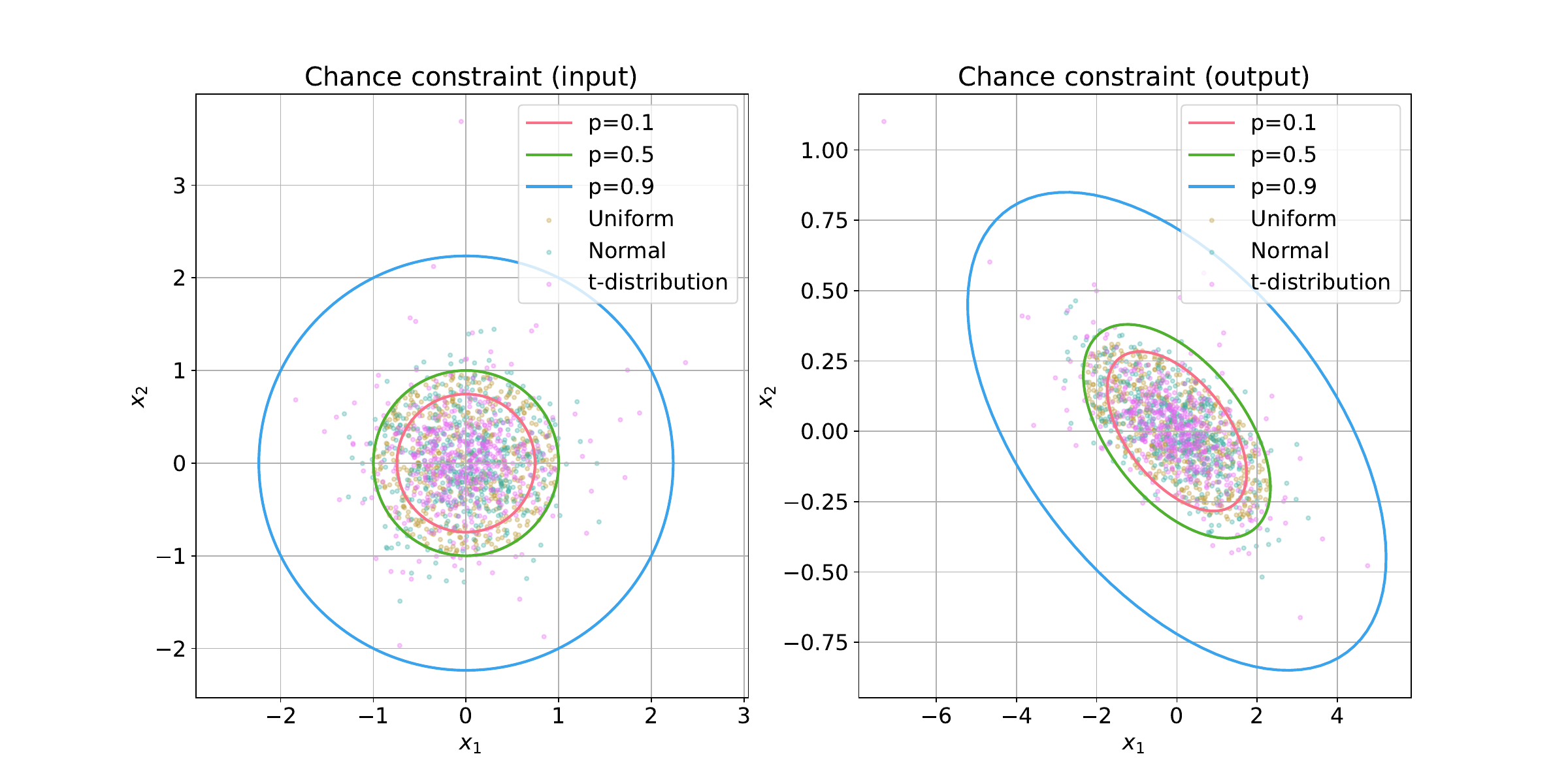} \\
    \subcaption{Confidence ellipsoid \cite{FazMP19}}
        \label{fig:chance}
    \end{subfigure}

    \vspace{5pt} 

    \begin{subfigure}{\linewidth}
        \centering
   \includegraphics[width=\linewidth, viewport=60 20 1045 530, clip]{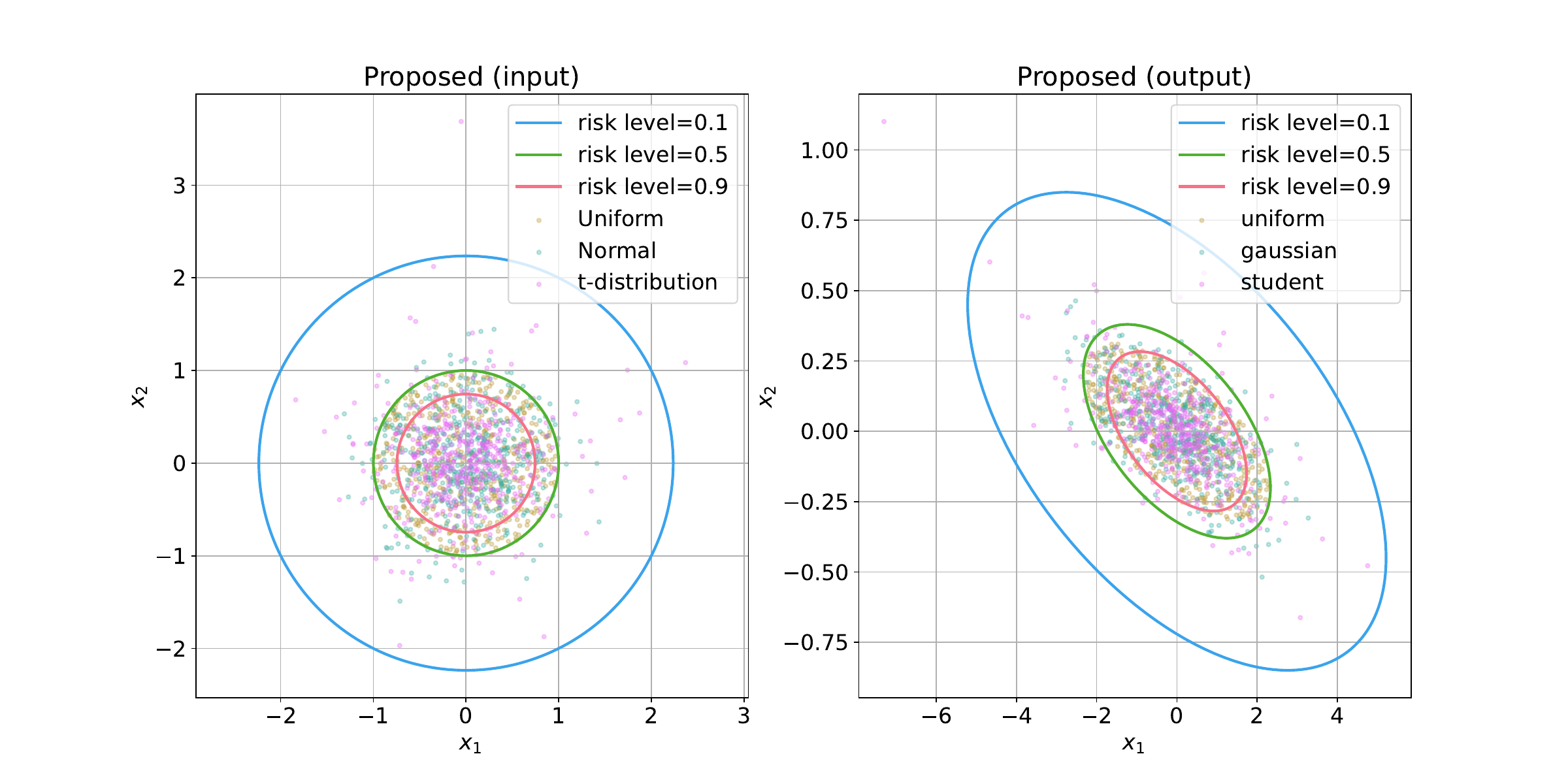} \\
    \subcaption{Risk-aware (proposed)}
        \label{fig:risk}
    \end{subfigure}
    
    \caption{Closed-loop reachability (left column: inputs, right column: outputs). Bounds are shown with  random samples $x$ in the input space and $x^+ =Ax+Bf(x)$ in the output space }
    \label{fig:reachability}
\end{figure}

Figures~\ref{fig:reachability}a)--c) illustrate the ellipsoidal safe sets obtained by the three approaches. 
The norm bounded approach ensures that all samples remain within the bound, as expected in Fig.~\ref{fig:reachability}(a). 
As expected, the confidence ellipsoid method (Fig.~\ref{fig:chance}) and the proposed risk-aware method (Fig.~\ref{fig:risk}) 
produce identical ellipsoids. We also observe that a $t$-distribution, which has a long tail, has samples out of those safe sets.

While this may seem to undermine its advantage, unlike \cite{FazMP19}, which is limited to ellipsoidal inputs, the proposed approach also applies to polytope and zonotope input, offering greater flexibility, such as classification problem we see the next.

%-----------------------------------
\subsection{Classifications}%classification_digits.ipynb

We consider a neural network classifier $f: \mathbb{R}^{n} \rightarrow \mathbb{R}^{m}$ that assigns input $x$ to the class with the highest score: $C(x) = \text{argmax}_{0 \leq i \leq m-1} f_i(x)$ \cite{HeiA17, FazMP22}. A classifier is risk-aware if the ambiguity set $\mathcal{P}$ does not alter the classification decision with respect to its mean $\mu$, i.e.,
for input $x\sim \mathbb{P} \in \mathcal{P}$, risk-aware classifier satisfies:
\begin{align} 
\sup_{\mathbb{P}\in \mathcal{P}}\mathbb{P}\text{-CVaR}_{\varepsilon}\left[f_c(x)-f_i(x)\right] \leq 0, \ \forall i \neq c.
\end{align}
Without loss of generality, we set $ c = C(\mu)=0$. Then, because  these constraints are in the form of polytope, using the results of \cite{FazMP22}, it can be written in the form of \eqref{eq:output} using 
\begin{align}\label{eq:class_S}
S = \begin{bmatrix} 0 & 0 & 0\\
 0 & S_{\text{sub}}^\top \Gamma S_{\text{sub}} & 0\\
 0 & 0 & 0
\end{bmatrix},\ S_{\text{sub}} = \begin{bmatrix}
 0 & 0\\
\mathbf{1}_{m-1} & -I_{m-1}
\end{bmatrix},
\end{align}
where $\Gamma\in\mathbb{S}^m$, $\Gamma \geq 0$, $\Gamma_{ii}=0$ with appropriate $S_{\text{sub}}$. 
Thus, the use of Theorem \ref{thm:sufficient} verifies the satisfaction of risk-aware classification.

We compare the performance of the proposed approach on different distributions, uniform, normal, Weibull, power law, lognormal, and student's $t$-distribution.
Experiments were performed using the scikit-learn Digits dataset (8×8 grayscale images of handwritten digits, 1797 samples) \cite{scikit-learn,AlpK98}.
A neural network classifier with 64 input neurons  (corresponding to the 8×8 input images), one hidden layer of 32  neurons, and 10 output neurons (one for each digit class 0-9, i.e., $f_i(x)$ is for digit class $i$ for $i = 0, \dots, 9$), which achieves ~98\% test accuracy was used. We set the risk-level $\varepsilon=0.2$. 

For robustness analysis, we first obtained $P$ for the risk-aware input QC such that $P$ along with $S$ as in \eqref{eq:class_S} and $Q$ in \eqref{eq:Q} of the neural network classifier satisfy the condition \eqref{eq:sum}. 
Then, we checked if this $P$  satisfies the risk-aware QC \eqref{eq:input}  with the mean and covariance of the images labeled as digit "6".
From Theorem \ref{thm:sufficient}, we expect the output satisfies \eqref{eq:output} if the test samples were generated from various distributions while preserving the mean and covariance of the images labeled as digit "6".

\begin{figure}[h]
    \centering

    % (a)
    \begin{subfigure}{\linewidth}
        \centering
    \includegraphics[width=\linewidth, viewport=60 40 980 508, clip]{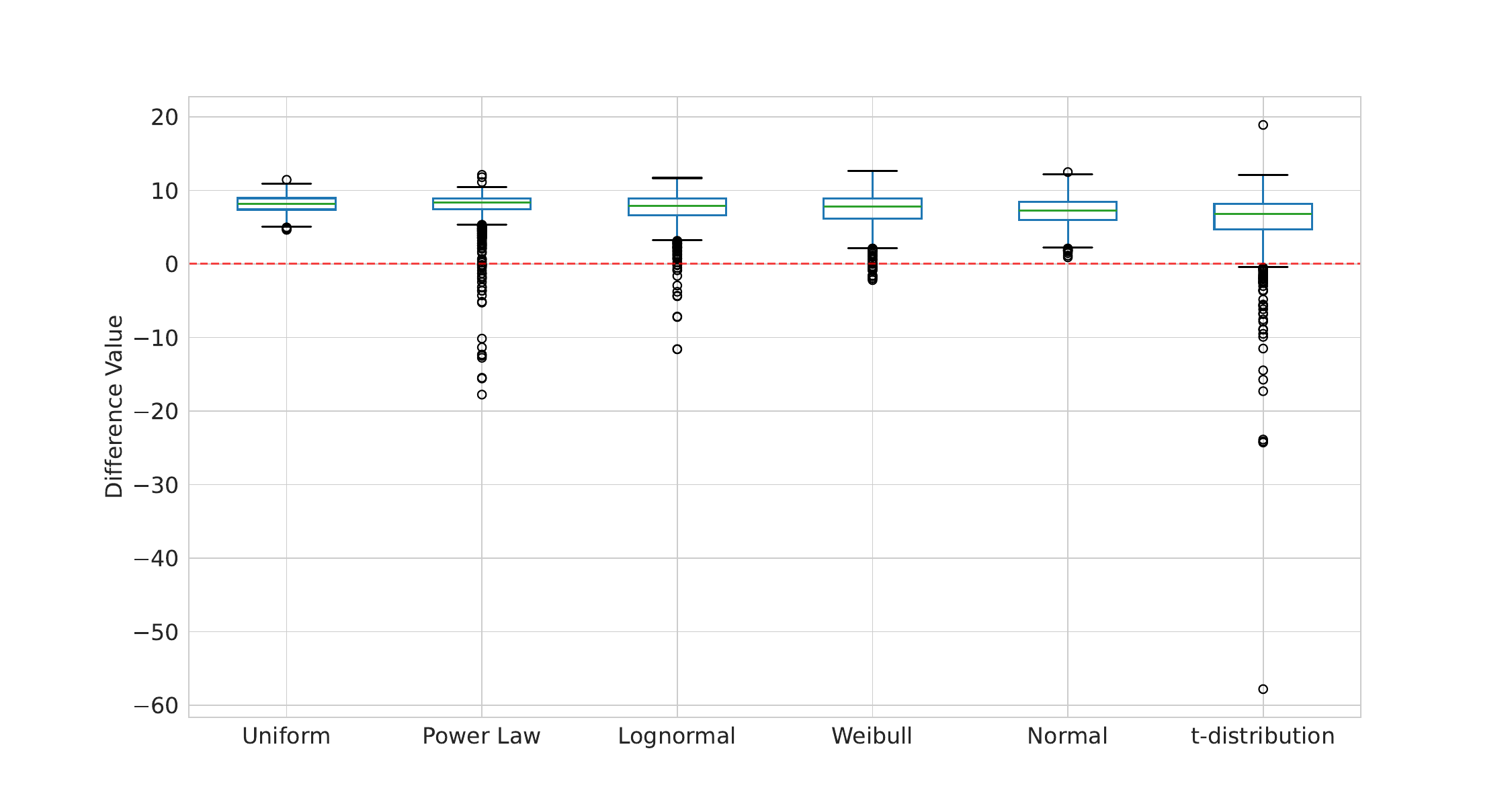}  \\
    \subcaption{Difference values, $f_6-\max_{i\neq 6} f_i$: A positive difference indicates correct classification}\label{fig:1}
        \label{fig:a}
    \end{subfigure}
    
    \vspace{5pt}

    % (b) & (c) 並列
    \begin{subfigure}{0.48\linewidth}
        \centering
        \includegraphics[width=\linewidth, viewport=100 20 370 300, clip]{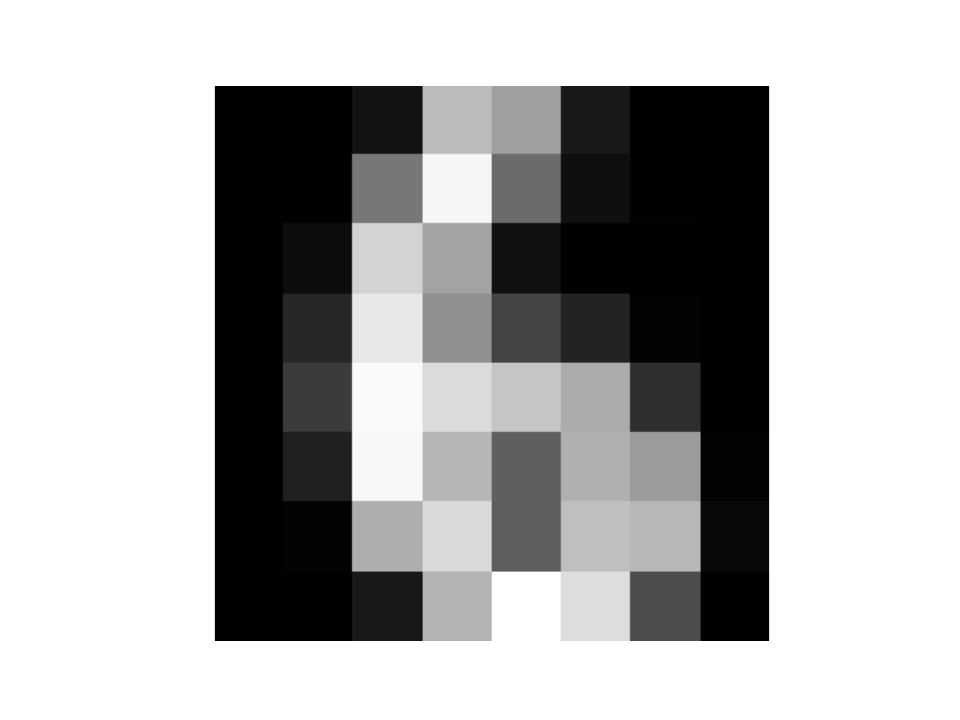}
        \caption{Mean of figures that are labeled `6'}
        \label{fig:b}
    \end{subfigure}
    \hfill
    \begin{subfigure}{0.48\linewidth}
        \centering
        \vspace{5pt} 
        \includegraphics[width=\linewidth, viewport=100 20 370 300, clip]{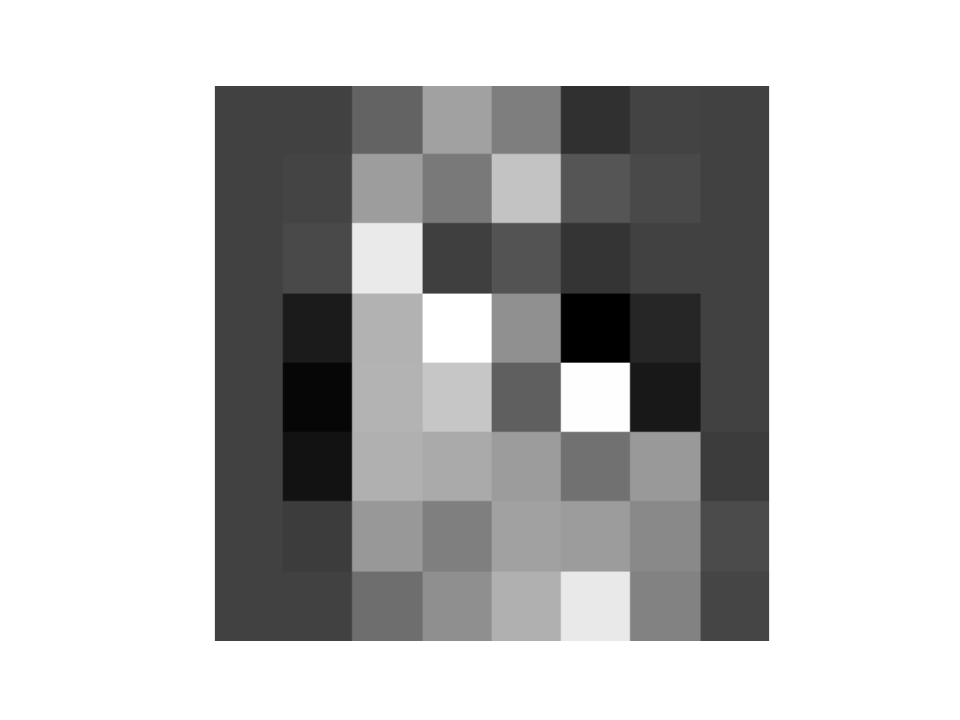} \vspace{-12pt} 
        \caption{A figure labeled '6' that was misclassified '8'. }
        \label{fig:c}
    \end{subfigure}

 % (d)
    \begin{subfigure}{\linewidth}
        \centering
   \includegraphics[width=\linewidth, viewport=50 20 650 320, clip]{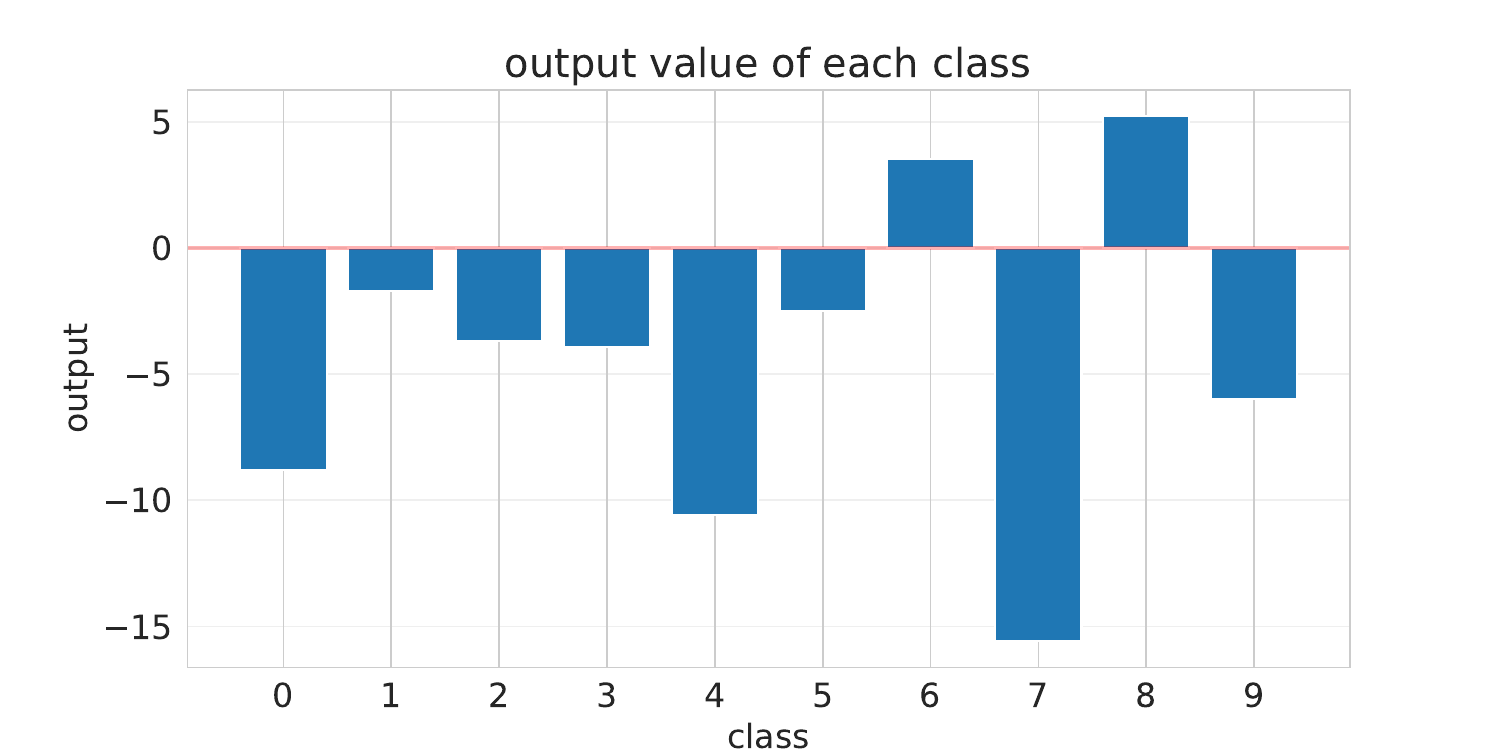} \\
    \subcaption{Output values of each class $f_i$ for misclassified sample in (c)}\label{fig:2}
        \label{fig:d}
    \end{subfigure}

    \caption{Classifications}
    \label{fig:digit}
\end{figure}

Figure~\ref{fig:digit}(a) shows the distribution of the difference values 
$f_6 - \max_{i \neq 6} f_i$ across various input distributions, 
where $f_6$ denotes the output for class ``6'' and $\max_{i \neq 6} f_i$ 
represents the maximum output among the other classes. 
A positive difference indicates correct classification, while a negative 
difference corresponds to a misclassification. The results highlight 
different levels of robustness: for instance, the uniform and normal distributions 
achieve perfect classification, whereas heavy-tailed distributions such as 
the $t$-distribution lead to more frequent misclassifications.

The statistical values of Fig. \ref{fig:digit}(a) are summarized in Table \ref{tab:diff_statistics}.
The mean/median columns show that uniform and power law have the largest typical margins, while heavy–tailed families exhibit visible skew (median $>$ mean), indicating occasional low or negative margins that pull the mean down; this skew is strongest for $t$-distribution. The standard deviation column separates stability: uniform distribution is the most concentrated (smallest spread), normal distribution is also tight, whereas  $t$-distribution is the most variable. The Positive Ratio (PR), defined as $\mathrm{PR}=\frac{1}{N}\sum_{k=1}^{N}\mathbf{1}\!\{\,P_{\text{diff}}(x^{(k)})>0\,\}$ with $P_{\text{diff}} := f_{\text{6}} - \max_{i \neq \text{6}} f_i$ and $x^{(k)}$ the $k$-th sample, then flags outright errors: all but student’s $t$-distribution are near or at $1$, i.e., misclassifications are rare for those inputs. Because PR saturates for several distributions, the decisive column is CVaR(0.20): uniform distribution sustains the largest positive tail margins (safest worst $20\%$), normal distribution remains safely positive, the other light-to-moderate tails (power law, lognormal, Weibull) are positive but weaker, and $t$-distribution alone turns negative—revealing tail failures despite acceptable central statistics.

Figure~\ref{fig:digit}(b) depicts the average feature pattern of samples correctly 
classified as ``6,'' capturing the typical shape of the digit. 
In contrast, Figure~\ref{fig:digit}(c) presents a misclassified ``6'' sample, 
where atypical handwriting leads the classifier to fail. 
Figure~\ref{fig:digit}(d) plots the output values $f_i$ for each class on this 
misclassified sample. It can be observed that both classes 6 and 8 yield 
positive activation values,  3.52 and 5.22, respectively, whereas the remaining classes have negative 
outputs. This indicates that the network confuses the digit ``6'' with 
``8,'' providing insight into the source of the misclassification, which aligns with human trends.

\begin{table*}[ht]
\centering
\caption{Summary statistics of $P_{\text{diff}}=f_{\text{6}}-\max_{i\neq \text{6}} f_i$ by input distribution.}
\label{tab:diff_statistics}
\begin{tabular}{lrrrrr}
\hline
Distribution & Mean & Median & Std.\ Dev. & Positive Ratio & CVaR (0.20) \\
\hline
Uniform       & 8.105 & 8.178 & 1.135 & 1.000 & 6.450 \\
Power law     & 7.637 & 8.364 & 2.797 & 0.973 & 3.941 \\
Lognormal     & 7.424 & 7.923 & 2.379 & 0.986 & 3.869 \\
Weibull       & 7.324 & 7.774 & 2.266 & 0.985 & 3.834 \\
Normal        & 7.116 & 7.210 & 1.927 & 1.000 & 4.280 \\
Student's $t$ & 5.841 & 6.803 & 4.344 & 0.943 & -0.092 \\
\hline
\end{tabular}
\end{table*}

%-----------------------------------
%%%%%%%%%%%%%
\section{Conclusions and Future Works} \label{sec:conc}
This paper introduced a risk-aware safety verification framework for neural networks under input uncertainties using WC-CVaR. 
Our approach provides risk-aware safety guarantees by controlling the expected severity of worst-case outcomes beyond a specified risk level. 
It allows us to balance conservatism with tail risk tolerance and to handle diverse input geometries beyond ellipsoids, while maintaining computational tractability at the same level as existing approaches. 
Experiments across reachability and classification tasks confirmed that our approach effectively handles heavy-tailed distributions. 

One possible future work is to broaden ambiguity sets beyond moments, and another would be to validate on safety-critical closed-loop systems while exploring controller synthesis with WC-CVaR.

%\addtolength{\textheight}{-1.2cm}   % This command serves to balance the column lengths
                                  % on the last page of the document manually. It shortens
                                  % the textheight of the last page by a suitable amount.
                                  % This command does not take effect until the next page
                                  % so it should come on the page before the last. Make
                                  % sure that you do not shorten the textheight too much.

%%%%%%%%%%%%%%%%%%%%%%%%%%%%%%%%%%%%%%%%%%%%%%%%%%%%%%%%%%%%%%%%%%%%%%%%%%%%%%%%
\section*{APPENDIX}
%------------------------------------------
\subsection{Expression of $Q$ for ReLU Function}
\label{appendix:Q} 

QC for ReLU is given in \cite{FazMP22} as follows:
The function $\phi(z) = \max(0,  z)$ satisfies the QC
\begin{align*}
\begin{bmatrix}
z\\
\phi(z) \\
1
\end{bmatrix}^{\top}
\begin{bmatrix}
Q_{11} & Q_{12} & Q_{13} \\
Q_{12}^{\top} & Q_{22} & Q_{23} \\
Q_{13}^{\top} & Q_{23}^{\top} & Q_{33}
\end{bmatrix}
\begin{bmatrix}
z \\
\phi(z) \\
1
\end{bmatrix} \geq 0
\end{align*}
for all $z \in \mathbb{R}^n$, where
\begin{align*}\begin{aligned}
Q_{11} &= 0, \quad Q_{12} = T,\quad
Q_{13} = -\nu, \\
 Q_{22} &= -2T, \quad
Q_{23} = \nu+\eta, \quad Q_{33} = 0,\\
T =&\sum_{i=1}^{n} \lambda_i e_i e_i^\top + \sum_{1 \leq i < j \leq n} \lambda_{ij} (e_i - e_j)(e_i - e_j)^{\top}, \\
& \lambda_{ij} \geq  0, \  \nu, \eta \geq 0.
\end{aligned}\end{align*}
Here, $e_i \in \mathbb{R}^n$ is the $i$th unit vector, $\lambda_i \in \mathbb{R}$, and 
 $\nu, \eta  \in \mathbb{R}^n$.

$\mathcal{Q}_{\phi}$ is the set of matrices $Q$ in this form.

%------------------------------------------
\subsection{Proof of Theorem \ref{thm:sufficient}} \label{proof}
We note that the theorem itself is essentially identical to Fazlyab’s and the computational cost to check the sufficient condition remains the same as in \cite{FazMP22}.
However, a new proof is required for our risk-aware case to incorporate WC-CVaR QCs. 

\begin{IEEEproof}
Let define a vector
$\bar{x} = [
x^{0^\top}\ x^{1^\top}\ \cdots \ x^{\ell^\top} \ 1
]^\top=[\mathbf{x}^\top \ 1]^\top$.
From \eqref{eq:sum}, it follows that 
\begin{align*}
-\bar{x} ^\top M_{\text{in}}(P)  \bar{x} -\underbrace{ \bar{x} ^\top M_{\text{mid}}(Q)  \bar{x} }_{\geq 0, \ \because  \eqref{eq:Q}}- \bar{x} ^\top M_{\text{out}}(S) \bar{x} \geq 0, 
\end{align*}
which implies
\begin{align*}
-\bar{x} ^\top M_{\text{in}}(P)  \bar{x} \geq \bar{x} ^\top M_{\text{out}}(S) \bar{x}.
\end{align*}
Proposition \ref{prop:sub} implies that  
\begin{align*}
&\underbrace{\sup_{\mathbb{P} \in \mathcal{P}} \mathbb{P}\text{-CVaR}_{\epsilon} \left[-\bar{x} ^\top M_{\text{in}}(P)  \bar{x}\right]}_{\leq 0, \ \because  \eqref{eq:input}}
\geq \sup_{\mathbb{P} \in \mathcal{P}} \mathbb{P}\text{-CVaR}_{\epsilon} \left[\bar{x} ^\top M_{\text{out}}(S) \bar{x}\right].
\end{align*}
Using
$
\bar{x} ^\top M_{\text{out}}(S) \bar{x}= \begin{bmatrix}
x \\ f(x) \\ 1
\end{bmatrix} ^\top S
\begin{bmatrix}
x \\ f(x) \\ 1
\end{bmatrix},
$
 we obtain the satisfaction of \eqref{eq:output}.
\end{IEEEproof}

%------------------------------------------
\subsection{Proof of Lemma \ref{lem:eq}}\label{appendix:equivalence} 
Let's see the intuition first.
Let $z=\Sigma^{-1/2}x$ and $Y=\|z\|^2$.
Then $\mathbb{E}[z]=0$, $\mathrm{Cov}(z)=I$, and $\mathbb{E}[Y]=n$.

Intuitively, among all distributions with $\mathbb{E}[z]=0$ and $\mathrm{Cov}(z)=I$,
both tail measures $\mathbb{E}[\mathbf 1\{Y>t\}]$ (where $\mathbf 1$ is an indicator) and $\mathbb{E}[(Y-\beta)^+]$
achieve their worst case when all probability mass is concentrated at just two radii:
some mass at the origin and the rest at one fixed radius $r$, i.e.,
\begin{align*}
\mathbb P(Y=r^2)=q,\qquad \mathbb P(Y=0)=1-q.
\end{align*}
Geometrically, a fraction $q$ of the mass lies on the sphere of radius $r$, and the remainder is at the origin.

Because both problems reduce to the same one–dimensional radius tradeoff,
the safe sets coincide once the ellipsoid’s shape is fixed and only its scale $k$.

\begin{IEEEproof}
From Lemma 2.4 in \cite{KisC23}, we have
\begin{align*}
\sup_{\mathbb{P}\in \mathcal{P}}\mathbb{P}\text{-CVaR}_{\varepsilon}[x^\top \Sigma^{-1}x] = \frac{1}{\varepsilon}\text{Tr}(\Sigma^{-1}  \Sigma)=\frac{n}{\varepsilon},
\end{align*}
recalling $n$ is the dimension of $x$.

On the other hand, Lemma 2 in \cite{FazMP19}, a $p$-level confidence region of $x$ is
\begin{align*}
\left\{x: x^\top \Sigma^{-1}x \leq \frac{n}{1-p} \right\}.
\end{align*}

Thus, the boundary of the risk-aware safe set with $\varepsilon$ is coincides with that of the confidence level set with $1-p$.
\end{IEEEproof}

This equivalence extends whenever one can define a variable $z$ with
$\mathbb{E}[z]=0$ and $\mathrm{Cov}(z)=I$, while fixing the ellipsoidal shape and tuning only its scale.

%%%%%%%%%%%%%%%%%%%%%%%%%%%%%%%%%%%%%%%%%%%%%%%%%%%%%%%%%%%%%%%%%%%%%%%%%%%%%%%%

\bibliographystyle{IEEEtran}
\bibliography{IEEEabrv,myref}

\end{document}